\documentclass{article}

\usepackage[nonatbib,preprint]{neurips_2023}

\usepackage[utf8]{inputenc} 
\usepackage[T1]{fontenc}    
\usepackage{hyperref}       
\usepackage{url}            
\usepackage{booktabs}       
\usepackage{amsfonts,amsmath, amsthm}       
\usepackage{nicefrac}       
\usepackage{microtype}      
\usepackage{xcolor}         
\usepackage{graphicx}       
\usepackage{multirow}  
\usepackage{cleveref} 
\usepackage{algorithm} 
\usepackage{adjustbox}
\usepackage{algpseudocode} 

\definecolor{colour1}{RGB}{128,100,227}
\definecolor{colour2}{RGB}{31,120,190}
\definecolor{colour3}{RGB}{178,55,250} 
\definecolor{colour4}{RGB}{51,160,44}
\definecolor{colour5}{RGB}{20,200,90}

\usepackage{xcolor, subfigure}

\theoremstyle{plain}

\newtoggle{editingMode}
\settoggle{editingMode}{true}

\newcommand{\SaGess}{\textsc{SaGess}} 

\newcommand{\DiGress}{\textsc{DiGress}}

\iftoggle{editingMode} 
{
    \newcounter{noteGLOBALctr} \setcounter{noteGLOBALctr}{1} 
    \newcommand{\comAE}[1]{\textbf{\textcolor{colour1}{{{Andrew: \#\arabic{noteGLOBALctr}: }}#1}} \addtocounter{noteGLOBALctr}{1}}

    \newcommand{\GR}[1]{\textbf{\textcolor{orange}{{{[Gesine:}}#1]}}}
     \newcommand{\MC}[1]{\textcolor{colour3}{{{MC: \#\arabic{noteGLOBALctr}: }}#1} \addtocounter{noteGLOBALctr}{1}}
    
     \newcommand{\SL}[1]{\textcolor{colour4}{{{Stratis: \#\arabic{noteGLOBALctr}: }}#1} \addtocounter{noteGLOBALctr}{1}}
    
    \newcommand{\CM}[1]{\textbf{\textcolor{cyan}{{{[Carsten: }}#1]}}}
}
{
     \newcommand{\comAE}[1]{}
     \newcommand{\GR}[1]{}
     \newcommand{\MC}[1]{}
     \newcommand{\SL}[1]{}
     \newcommand{\CM}[1]{}
}

\newcommand{\close}[1]{\underline{\textcolor{blue}{#1}}}
\newcommand{\twoclose}[1]{\textcolor{colour4}{#1}}

\newcommand{\gr}[1]{{\color{magenta} #1}}

\title{SaGess: Sampling Graph Denoising Diffusion Model for Scalable Graph Generation}

\newcommand{\ER}{\text{Erd\H{o}s-R\'{e}nyi}}

\author{Stratis Limnios \\
The Alan Turing Institute\\
London, UK\\
\texttt{slimnios@turing.ac.uk}\\
\And
\textbf{Praveen Selvaraj} \\
The Alan Turing Institute \\
London, UK\\
\texttt{pselvaraj@turing.ac.uk}
\And
\textbf{Mihai Cucuringu}\\
  Department of Statistics \& Mathematical Institute\\
  University of Oxford \&  \\
  The Alan Turing Institute,  London, UK \\ 
  \texttt{mihai.cucuringu@stats.ox.ac.uk} \\
 \And
 \textbf{Carsten Maple}\\
 University of Warwick\\
 Coventry, UK\\
 \texttt{cm@warwick.ac.uk}\\
\And
  \textbf{Gesine Reinert} \\ 
  Department of Statistics\\
  University of Oxford  \&  \\ 
  The Alan Turing Institute,  London, UK \\
  \texttt{reinert@stats.ox.ac.uk} \\
\And
  \textbf{Andrew Elliott}\\
  Department of Mathematics and Statistics\\
  University of Glasgow \&  \\
  The Alan Turing Institute,  London, UK \\ 
  \texttt{ Andrew.Elliott@glasgow.ac.uk} \\
}

\definecolor{colour3}{RGB}{178,55,250} 

\begin{document}

\maketitle

\begin{abstract}
Over recent years, denoising diffusion generative models have come to be considered as state-of-the-art methods for synthetic data generation, especially in the case of generating images. These approaches have also proved successful in other applications such as tabular and graph data generation. However, due to computational complexity, to this date, the application of these techniques to graph data has been restricted to small graphs, such as those used in molecular modeling.  In this paper, we propose {\SaGess}, a discrete denoising diffusion approach, which is able 
to generate large real-world networks by augmenting a diffusion model ({\DiGress}) with a generalized divide-and-conquer framework.
The algorithm is capable of generating larger graphs by sampling a covering of subgraphs of the initial graph in order to train {\DiGress}. {\SaGess} then constructs a synthetic graph using the subgraphs that have been generated by {\DiGress}. We evaluate the quality of the synthetic data sets against several competitor methods by comparing graph statistics between the original and synthetic samples, as well as evaluating the utility of the synthetic data set produced by using it to train a task-driven model, namely link prediction. In our experiments, \SaGess\, outperforms most of the one-shot state-of-the-art graph generating methods  by a significant factor, both on the graph metrics and on the link prediction task.
\end{abstract}

\section{Introduction}


Synthetic data are key to many methods in machine learning and statistics, and synthetic data generation has sparked a significant amount of attention in recent years. Tools such as Dall-E for synthetic image generation, and ChatGPT  have triggered the curiosity of the general public; the majority of the machine learning community are also diverting interest and resources towards generative algorithms. Beyond generating appealing synthetic images or asking ChatGPT to write prose, many real world applications benefit greatly from synthetic data generation, for tasks including data augmentation in the training of classifiers/regressors\cite{wong2016understanding}, privacy protection of sensitive data\cite{beaulieu2019privacy} or removing bias from data sets \cite{van2021decaf}. Synthetic graph generation for modeling social interactions, generating new chemical compounds, or forecasting transactions are capital tasks that require efficient methods. 

Many complex data sets can be represented as networks, and hence synthetic graph generators are of particular interest. Often these networks are viewed as realizations of a random process. The design of generative models for random graphs has 
a rich history, coming from probabilistic and structural assumptions with traditional 
methods such as early work in \ER graphs \cite{erdHos1960evolution}, stochastic block models \cite{HOLLAND1983109}, exponential random graphs \cite{ROBINS2007173} or the Barabasi-Albert model \cite{RevModPhys.74.47}. 
However, these methods often oversimplify the underlying complex structure of the graphs and are often not able to 
capture the distributions arising from real-world scenarios. Thus, more recently there has been an increasing interest in the community in designing deep models for synthetic graph generation,  which allows for more flexible algorithms that are able to capture the intricacies of real networks with complex dependencies between the edges.  
Many approaches have been proposed for graph generation
 including autoregressive approaches which generate nodes and edges step by step. Such methods include GraphRNN \cite{you2018graphrnn} and GRAN \cite{liao2019gran}, 
 which improves modeling of long-term dependencies using a graph-based attention mechanism. Additional approaches include 
 autoencoder based approaches, such as GraphVAE \cite{simonovsky2018graphvae}, adversarial approaches such as ~\cite{bojchevski2018netgan}.
There are many more methods; detailed reviews are found for example in~\cite{faez2021deep,guo2022systematic}.

Moreover, after the outstanding performances of Denoising Diffusion Models \cite{ho2020denoising,sohl2015deep} on image generation, 
various lines of work have applied such models to graph learning, mostly for molecular generation such as GeoDiff \cite{xu2022geodiff} or chemical compound design \cite{schneuing2022structure}. More versatile models have been introduced, such as {\DiGress} \cite{vignac2022digress}, a discrete denoising diffusion graph model that is able to generate very high quality graphs with node and edge attributes. One of the main assets of denoising diffusion models is not having to rely on adversarial training, but they still need a large data set to train. The implication of this is that models such as {\DiGress} do not adapt well when the task is to generate one large graph from a single sample.

Formally, creating a synthetic graph generator is equivalent, either explicitly or implicitly, to estimating/sampling from a probability distribution over space of possibly directed, possibly weighted/attributed graphs.  When many independent realizations from the unknown distribution are available, then standard estimation methods are often successful.
%
%
Yet, many real world graph data sets consist of only a single graph, either because it is expensive to measure, or because of the nature of the data, for example  a global social network. Thus, there is interest in the 
more difficult task of approximating 
the underlying probability distribution only viewing a single sample. Classical models often navigate this task by making  strong assumptions,
such as {\ER}, where we assume the edges are i.i.d..
Recent deep learning methods often approach this challenge by the structured model of a GAE/GVAE
or by learning the distribution over subsamples of nodes and edges; an example is 
\footnote{In effect, this assumes that the underlying process that generated the network is related to these samples.}
e.g. NetGAN, an adversarial approach which relies on random walks~\cite{bojchevski2018netgan}.

In particular, while {\DiGress} is a powerful approach for generating small graphs, its requirement to have multiple training samples as input prevents its application to the situation when the input is just a single network. Building on  {\DiGress},
in this work we propose a denoising-based diffusion model that can operate in the single large graph data set case. To achieve this, we  leverage the strengths and the quality of the synthetic {graph} data produced by DiGress, in order to design a denoising diffusion based model.  Along the way, we alleviate two of its weaknesses in this context, not only its requirement for a large number of training samples, which are not available in this case, but also its poor scalability to larger graphs.

We achieve this by leveraging a divide-and-conquer scheme, which allows us to extend this method to larger {single} graphs.

Our main contributions can be summarized as follows. 

{$\bullet$} 
   We introduce graph subsampling methods, to break large graphs into a trainable data set.

{$\bullet$} We propose {\SaGess} (SAmpling Graph dEnoiSing DiffuSion model), a pipeline that allows us to employ {\DiGress} as a sample generating base for larger graphs.

{$\bullet$} We propose a task-driven evaluation by training link prediction GVAE on synthetic data and testing on real.

The paper is structured as follows. In Section \ref{sec:Digress} we set up the notations for the rest of the paper, as well as the {\DiGress} model that is at the center of our framework, while also elaborating on its limitations. In Section \ref{sec:sagess}, we state the problem we aim to address, and then present our solution framework {\SaGess}, which unfolds in two sections: first, the sampling methods to obtain a training data set, and second, the reconstruction pipeline of the synthetic graph. Next, in sections \ref{sec:experiments} and \ref{sec:results}, we respectively present our experimental setup and results. Finally, we discuss potential limitations and future work in Section  \ref{sec:concl}.

\section{Graph Diffusion Model and Scalability}
\label{sec:Digress}

Notations and definitions presented in this section will be used throughout the rest of the paper. We  provide the essential framework to support our work,  and elaborate on the problems we address with the introduction of the {\SaGess} framework. 

Graphs in this paper are denoted as $G=(V,E)$ where $|V|=n$ is the set of nodes and $|E|=e$ is the set of edges. Each node is of one of $a$ types, and each edge is of one of $b$ types. We associate with $G$ the matrices $\mathbf{X}\in \mathbf{R}^{n\times a }$ where $\mathbf{X}_{i,:}$ is the one-hot encoding for the feature of node $i$, and  $\mathbf{E}\in \mathbf{R}^{n\times n \times b}$ where $\mathbf{E}_{i,j,:}$ is the one-hot encoding for the feature of the edge between nodes $i$ and $j$.
We denote by $P_k(V)$ the set of all $k$-point subsets of $V$ and by 
$\mathcal{S}_k(G)$ the set of all possible subgraphs 
of $G$ of size $k$. The subgraph of $G$ induced by the nodes in $S$ is denoted by $G[S]$; we also refer to such graphs as {\it  node induced subgraphs}.

\subsection{Graph Generation using Discrete Denoising Diffusion Model:  {\DiGress}}

Here we introduce the key aspects of Denoising Diffusion Probabilistic Models (DDPM) with particular emphasis on application 
to graphs. We also point out limitations of state-of-the art graph generation using discrete space diffusion models. 

{\DiGress} \cite{vignac2022digress} is currently one of the most efficient tools in graph generation. Taking as input a data set of a variety of graphs, it learns a denoising process in discrete space and is able to mimic the input graphs with remarkable precision. The key aspect of this method, and more generally graph generation methods based on diffusion models \cite{haefeli2022diffusion,vignac2022digress}, relies on a discrete space noise scheduling. Indeed, instead of the standard Gaussian noising and denoising procedure, these frameworks propose to add iteratively discrete noise via edge and node addition and deletion at random. 

Traditionally, these models are based on a forward and a reverse Markov process. Indeed a forward process denoted by $q(A^{1:T}|A_0) = \prod_{1=t}^T q(A^{t}|A^{t-1})$ generates an increasingly noisier samples from the candidate $A^0$ to white noise $A^T$, $A^i$ here being adjacency matrices and $T$ is a hyperparameter; in {\DiGress} the default is $500$.  We then learn the reverse process $p_{\theta}(A^{1:T}) = p(A^T)\prod_{1=t}^T q(A^{t-1}|A^{t})$ that aims to denoise the latent adjacency matrices $A^t$ to produce the synthetic samples.

The {\DiGress} model takes as state space the set of node types and of edge types. {\DiGress} then defines the transition probabilities from one state to another for nodes and edges through the noise matrices $[Q_X]_{i,j}^t = q(x^t=j| x^{t-1} =i)$ and $[Q_E]_{i,j}^t = q(e^t=j| e^{t-1} =i)$ where $Q$ is chosen so that the Markov chain converges to the relative type frequencies in the initial population. Then, to get a noisy sample $G^t= (\mathbf{X}^t,\mathbf{E}^t)$ each node and each edge type is sampled from the categorical distribution 
$q(G^t|G)= (\mathbf{X}\bar{\mathbf{Q}}^t_X,\mathbf{E}\bar{\mathbf{Q}}^t_E)$ with $\bar{\mathbf{Q}}^t_X = \mathbf{Q}^1_X \dots \mathbf{Q}^t_X $ and $\bar{\mathbf{Q}}^t_E = \mathbf{Q}^1_E \dots \mathbf{Q}^t_E $.

The denoising component of the {\DiGress} model is a denoising neural network $\phi_{\theta}$ parametrized by $\theta$. It is trained by optimizing the cross-entropy loss between the predicted probabilities $\hat{p} = ({\hat p}^X, {\hat p}^E) $ and the true graph $G$, and is defined by
\begin{equation}
    l(\hat{p}^G,G) = \sum_{1\le i \le n} \textbf{cross-entropy}
    (x_i,\hat{p}_i^X) + \lambda \sum_{1\le i \le n} \textbf{cross-entropy}(e_{ij},\hat{p}_{ij}^E), 
\end{equation}
where $\lambda \in \mathbb{R}^+$ balances the importance between nodes and edges.

Simply put, {\DiGress} is a way to obtain a classification for each node and  edge.

Finally, after training the network, it can be used to generate synthetic graphs via the estimation of the reverse diffusion iteration $p_{\theta}(G^{t}|G)$ using $\hat{p}^G$, as follows. We put $p_\theta
(x_i^{t-1} | x_i=x, G^t) = q(x_i^{t-1}| x_i=x, x_i^t) {\mathbf{1}}( q( x_i^t | x_i=x) > 0) $ and $p_\theta
(e_{ij}^{t-1} |e_{ij}=e, G^t) = q(e_{ij}^{t-1}| e_{ij}=e, e_{ij}^t) {\mathbf{1}}( q( e_{ij}^t | e_{ij}=e) > 0) $, then set
\[p_\theta
(x_i^{t-1} | G^t) = 
\sum_x p_\theta
(x_i^{t-1} | x_i=x, G^t) {\hat p}_i^X (x),  \quad 
p_\theta
(e_{ij}^{t-1} | G^t) = 
\sum_x p_\theta
(e_{ij}^{t-1} | e_{ij}=e, G^t) {\hat p}_{ij}^E (e), 
\]
and finally obtain
$ p_\theta (G^{t-1}| G^t)  = \prod_{i=1}^n p_\theta
(x_i^{t-1} | G^t) \prod_{1 \le i,j \le n} p_\theta
(e_{ij}^{t-1} | G^t).$
The sampled $G^{t-1}$ then serves as input of the denoising network at the next time step.
This derivation shows the complexity of the calculations; when the numbers $a$ and $b$ of types are small, when the network is small, and when $T$ is small, they are feasible, but large networks with many different types can pose a challenge. More details are in the SI.

Overall, {\DiGress} can perform extremely well to generate small graphs with attributes from a finite set of attributes. Unfortunately, the complexity of {\DiGress} and the way the model is set up {in a way that prevents us from using it} to sample one large graph. 
The complexity  lies in the dimensions of the matrices $\mathbf{X},\bar{\mathbf{Q}}^t_X,\mathbf{E}$ and $\bar{\mathbf{Q}}^t_E$ as at least the edge matrices scale in $n^2$ times the number of attributes, it is indeed specified in the \cite{vignac2022digress} that {\DiGress} has complexity $\Theta(n^2)$ per layer, due to the attention scores and the predictions for each edge. Hence, not only we need to store in memory these matrices for every diffusion step, but we also need to learn $T$ dense matrices $Q_E$ to define the Markov transitions. This is fairly reasonable for small graphs, but starts to become computationally difficult to approach when the number of nodes in the graph increases. Then,  supposing we are able to run {\DiGress} on large graphs, we cannot train it on one single graph. {\DiGress}, much like other deep-learning algorithms, thrives when having access to a large training set. Altogether, this provides the motivation for proposing  our framework based on sampling from large graphs.

\section{ Sampling Graph Denoising Diffusion Model Generator ({\SaGess})}
\label{sec:sagess}

\begin{figure}[t]
    \centering
    \includegraphics[width=\textwidth]{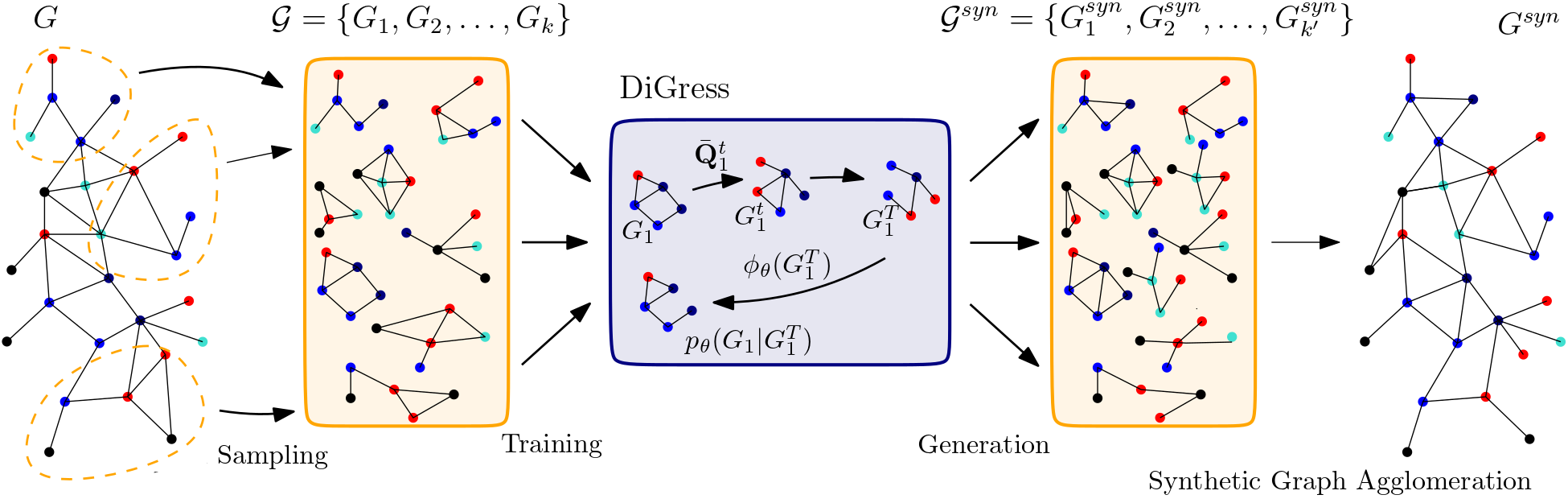}
    \caption{{\SaGess} Diagram } 
    \label{fig:sagess_diag}
\end{figure}

This paper addresses the task to generate a single graph from a unique observation. We do not have a large training set of graphs $\mathcal{G}$ available to train any type of diffusion model. Our solution, which we denote as {\SaGess} and which is 
 visualized in Figure \ref{fig:sagess_diag}, can be summarized as follows.

1. First, we produce a graph data set $\mathcal{G}$ issued from the initial graph $G$.

2. Next, we employ {\DiGress} to create samples using the data set $\mathcal{G}$ as training data.

3. Finally, we rebuild a graph from the trained diffusion model in a systematic fashion, paying  particular care to match the node ids in the samples.

\subsection{Graph Sampling and Covering}

The first step is to construct a representative data set to train {\DiGress}.
We propose three sampling schemes to capture different structural properties of the graph. Each method addresses one or more properties of the graph, from local to more global ones. Depending on the real world graph we apply our model to, some  sampling schemes might be more effective than others; for instance, ego networks might benefit more from locality in the sampling, whereas communication data sets, for instance, might benefit more from more global features such as random walks.

\paragraph{Uniform Node Sampling (\textbf{Unif}):}
We sample a set of node induced subgraphs by selecting $k$ sized subsets of nodes $S_k $  uniformly at random from $P_k(V)$. We note that this sampling scheme is  invariant under permutation of the nodes. Based on results from \cite{mcgregor_et_al:LIPIcs.ICALP.2022.96},
if $p$ denotes the probability of selecting a node, then 
$\Theta (p^{-2} \log n \log(\nicefrac{1}{\delta}))$ uniformly node induced subgraphs suffice  for reconstruction with probability at least $1-\delta$. In our setting, $p = k/n$. To give a broad sense to the number of subgraphs we need to sample, for a graph with $n = 1,000$ nodes, suppose we select uniformly at random $2\log(n)$ nodes. We would then need $\Theta(4\log^3(n) \log(\nicefrac{1}{\delta}))$ subgraphs of size $20$ to reconstruct $G$. Setting the probability $1 - \delta =  0.95$ results in a data set size of the order of $10,000$ graphs. Here we use $10,000$ for simplicity; other values are explored in the SI.

\paragraph{Random Walk Node Induced Subgraph (\textbf{RW}):}

This second method is based on sampling from subgraphs induced by random walks starting from every node in the graph $G$. The idea is straightforward: for each node $v_1$ of the $n$ nodes in $V$ we generate a  $k$ step random walk $w^{k}=\{v_1,\dots,v_k\}$ starting from $v_1$.
Then we obtain the node induced subgraph from those walks $G[w_k]$. We repeat this construction $d$ times. This gives us a data set $\mathcal{G} = \{G[w^k_1],\dots,G[w^k_{d \times n}]\}$, where $d$ is a hyper-parameter controlling the number of sampled graphs per node, see supplementary material for additional discussion.

\paragraph{$2$-hop Neighborhood Sampling (\textbf{Ego}):}

The last sampling method is based on sampling from every node a $2$-hop neighborhood with random node deletion.
As a $2$-hop neighborhood can cover most of the graph and we aim for small samples to be able to train {\DiGress}, with ideally more than one subgraph including each node, we fix to $k$ the maximum size of the subgraph we generate from the $2$-hop neighborhood of a node. To do so, if the neighborhood is larger than $k$,  we delete (the integer part of) half of nodes from the neighborhood uniformly at random.
We also ensure to keep, at every step, the largest connected component until 
there are less than $k$ nodes left; we denote this resulting node set by 

$\mathcal{N}^k(v)$. 
Again, we obtain its node induced subgraph $ G[\mathcal{N}^k(v)]$. 
For $k$ small enough, there is randomness in our modified $2$-hop neighborhoods. 
We generate $d$ subgraphs per node, with $d$ 
as in the previous paragraph,  hence producing a data set $\mathcal{G}:=\{ G[\mathcal{N}^k_1(v_1)],\dots,G[\mathcal{N}^k_d(v_1)],\dots,G[\mathcal{N}^k_1(v_n)],\dots,G[\mathcal{N}^k_d(v_n)]\}$.
Theoretical guarantees for this subsampling scheme are available for example in \cite{ali2016comparison}.

\begin{algorithm}
	\caption{{Ego Sampling}} 
	\begin{algorithmic}[1]
    \For {$i \in \{1,\dots,n\}$}  
        \For {$j \in \{1,\dots,d\}$}
        \State $G_i = \mathcal{N}(v_i)$
        \While {$|V_i| >k$}
            \State choose uniformly $\lfloor \frac{|V_i|}{2} \rfloor $ nodes from $V_i$; call this set  $S_i$
            \State delete the nodes in $S_i$ from $G_i$
            \State retain the largest connected component of $G_i$
        \EndWhile
        \EndFor
    \EndFor
	\end{algorithmic} 
\end{algorithm}

Each of these methods provides a graph training set for {\DiGress} which is obtained from one single graph. Moreover, following \cite{mcgregor_et_al:LIPIcs.ICALP.2022.96}, we produce enough samples to have a good chance to cover each node and edge in the uniform sampling scheme, with the heuristic that this should also be 
adequate for the denser sampling schemes. We also shuffle the training data set $\mathcal{G}$ before feeding it to {\DiGress}.
Next, we detail how {\SaGess} reconstructs the initial graph from the trained model.

\subsection{Node Labeling and Reconstruction}

Our {\SaGess} model first samples subgraphs to train {\DiGress}, but {\DiGress} will only be able to generate small graphs from the training set $\mathcal{G}$ it learns from. This would lead to a non-identifiable set of small graphs, that one would not know how to merge into one single graph.

This is where the ability of {\DiGress} to handle node attributes is important. We set the initial node ids as node features on the subgraphs 
to enable identification after generating synthetic samples. In detail, to every graph $G^k_i \in\mathcal{G}$ we associate a feature matrix ${\bf X_i} \in \mathbb{R}^{k \times n}$, where $k$ is the number of nodes in $G^k_i$ and $n$ the number of nodes in $G$, so that ${\bf X_i}$ is a one-hot encoding of the node id in the initial graph. 
 This enables the model to learn the local graph structure as we identify the nodes with the attribute. We then generate enough small graph samples with {\DiGress} and agglomerate the new edges and nodes uniquely until we match the initial graph $G$'s edge count. 
 The agglomeration step unfolds as follows. We start with the first generated graph. Then as long as the number of edges, $|E|$, in the original graph is not reached, we 
 generate a new synthetic graph $G_{syn}^k$ with {\DiGress} and take the graph union with the current graph.
 The sampling process stop when $|E_{syn}|>|E|$; 
 if $G_{syn}^l$ is the last graph sampled, we add all the new edges to $G_{syn}$ regardless of 
 whether the new edge count exceeds
 $|E_{syn}|$. As the generated subgraphs tend to be small, the overshoot tends to be small also.

\begin{algorithm}
	\caption{{\SaGess}} 
	\begin{algorithmic}[1]
     \State $\mathcal{G}$: Sample $n\times d$ samples using Unif/Ego/RW from $G$
	\State Train {\DiGress} with $\mathcal{G}$	
    \While {$|E_{syn}| < |E|$}
        \State generate $G^{k}_{syn}$
        \State add new unique edges from $(G^{k}_{syn})$ to $G_{syn}$ 
    \EndWhile
	\end{algorithmic} 
\end{algorithm}

As it builds on {\DiGress}, {\SaGess} comes with theoretical guarantees inherited from {\DiGress}, in particular the permutation equivariant architecture and a permutation invariant loss. Exchangeability of the generated distributions then follows under the exchangeable {\SaGess} sampling schemes, permuting the order of the samples in the input to destroy any potential sequential dependence. These properties ensure that {\SaGess} can learn efficiently from the data.

\section{Experimental Evaluation}
\label{sec:experiments}

In this section we will be evaluating our framework against state-of-the-art graph generation methods on four real world data sets and one synthetic data set. First, we will compare graph statistics to evaluate the quality of the generated graphs. Then we will train a Variational Graph Auto-Encoder (GVAE) to evaluate the utility of the synthetic data generated on a link prediction task.

\subsection{Benchmark models}
To benchmark our approach, we will compare against several competitor methods which construct synthetic data based on a single sample. To make a meaningful comparison, we have selected methods from multiple different generation approaches, from classical approaches, through adversarial approaches, and approaches based on a low-rank approximation. 
They are as follows:

{\bf DCSBM \cite{karrer2011stochastic}} 
A classic approach from the network science literature, the so-called degree corrected stochastic block model\footnote{Implementation from 
\url{https://github.com/microsoft/graspologic}} assumes a classic stochastic block model, where the nodes are divided into blocks, and the probability of a connection between a pair of nodes is a function of their block membership and additionally their degree. 
\\
{\bf NetGAN \cite{bojchevski2018netgan}}
An adversarial approach leveraging the GAN framework, to learn the distribution of biased random walks which are then combined via a transition matrix into a sampled graph. Edges in the sampled graph are sampled (mostly) uniformly at random. 
\footnote{Implementation from \url{https://github.com/danielzuegner/netgan}}
\\
{\bf CELL \cite{pmlr-v119-rendsburg20a}}
A modification of NetGAN which replaces the GAN formulation with a formulation based on a low-rank approximation, see paper for full discussion. In this formulation, sampled graph edges are again sampled in an edge independent manner.\footnote{Implementation from 
\url{https://github.com/hheidrich/CELL}}

\subsection{Data sets}

We evaluate our method on four real world data sets from the torch geometric package \footnote{\url{https://pytorch-geometric.readthedocs.io/en/latest/modules/datasets.html}} and one synthetic data set:

\textbf{EuCore:} This data set is an e-mail communication network of a large European research institution, from \cite{eucore}. Nodes indicate members of the institution, and an edge between a pair of members indicates that they exchanged at least one email. This graph has $1005$ nodes and $16,706$ edges.\\
\textbf{Cora:} Cora is a citation data set from \cite{yang2016revisiting}. Every node is an article and an edge links two nodes if one cites the other. It consists of a directed graph with $2,708$ nodes and $10,556$ edges. \\
\textbf{Wiki:} A data set from wikipedia pages from \cite{yang2020scaling} with $2,405$ nodes and $12,761$ undirected edges. \\
\textbf{Facebook:} This data set consists of friend lists from Facebook published in \cite{facebook}. This data set initially contains 10 graphs, but we will only be using the largest one (second graph) which has $1,045$ nodes and $27,755$ undirected edges.\\
\textbf{SBM:} This is a standard Stochastic Block Model (SBM) generated graph, with $4$ blocks of sizes $400$. We set the inner-cluster density to $0.15$ and the across-cluster density to $0.01$.

\subsection{Experiments}

\begin{table}[t] 
\caption{Comparing network statistics of graphs generated from each synthetic data method on five different graph data sets. This table includes the graph statistics  of Table 2 in Ref.~\cite{pmlr-v119-rendsburg20a}, from a similar experiment, with CPL denoting the characteristic path length; we also report the number of nodes and the number of edges. 
The text \close{ABC} in blue indicates the closest statistic to the real data, whereas the green text  \twoclose{ABC} indicates the second closest. 
}

\label{tab:table1}
    \centering

\begin{adjustbox}{width=\textwidth}
    \begin{tabular}{ccccccccccc}
    \toprule
\multirow{2}{*}{ }  &  
\multirow{2}{*}{Method}          & 
 num              &   
 num              &  
 num              & 
 num              & 
 max              &
 cluster          &
 \multirow{2}{*}{assort.}    & 
 power            & 
  \multirow{2}{*}{CPL} \\ 
                  &  
                  & 
 nodes            &   
 edges            &  
     triangles    & 
     squares      & 
 deg              &
            coef. &
                  & 
        law exp    & 
     \\ 

    \midrule

\multirow{ 7}{*}{\rotatebox{90}{EuCore}}
& Real    &   $1005$ &  $16,706$ &  $105,461$ & $4,939,311$ & $346 $ & $0.39935$ & $-0.01099$ & $1.3621$ & $2.58693$\\ 
                         & {\SaGess}-Uni     &   $939$ &  $16,716$ & \twoclose{$114,900$} & $6,280,664$ & $287$  &  $0.34024$ &  $-0.06321$ & \twoclose{$1.35697$}& $ 2.49139$ \\
                         & {\SaGess}-RW &   $878$ &  $16,709$ &  $131,429$ & $6,995,335$ & \close{$351$} &  \twoclose{$0.42338$} &  $-0.04234$ & $1.35192$& \twoclose{$2.50512$} \\ 
                        & {\SaGess}-Ego    &   $867$ &  $16,707$ &  \close{$114,593$} & \close{$5,257,738$} & \twoclose{$342$} &  \close{$0.39000$} &                  \twoclose{$-0.023969$} & $1.3296$ & $2.43646$\\
                       
                       & NetGAN    &   $986$ &  $16,064$ &  $62,278$  & $2,505,330$ & $279$ & $0.25569$ &  
                       $-0.06196$ & $1.34179$ & $2.48189$
                       \\  &  CELL    &   $1005$ &  $16,064$ &  $74,251$ & $3,336,294$ & $273$ &  $0.29808$ &  
                       $-0.07655$ & \close{$1.36279$} & \close{$2.56782$}
                       \\  &  DCSBM    &   $951$ &  $15,906$ &  $75,743$ & \twoclose{$3,699,308$} & $305$ &  $0.19673$ & 
                       \close{$-0.010515$} & $1.35087$ & $2.47134$
                        \\ \midrule
\multirow{ 6}{*}{\rotatebox{90}{Cora}}
& Real    &   $2708$ &  $10,556$ &  $1,630$  & $4,664$ & $168$ & $0.24067$ &  $-0.06587$ & $1.93230$ & $6.31031$\\ 
                        & {\SaGess}-RW  &   $2548$ &  $10,557$ & \twoclose{$1,806$} & $14,952$ & $220$ &  \close{$0.25728$} &  \twoclose{$-0.05969$} & $1.84290$ & $5.43069 $\\ 
                        & {\SaGess}-Ego &   $2540$ &  $10,562$ & \close{$ 1,696$} & $7,923$ & \close{$171$} &  \close{$0.23911$} &  \close{$-0.06868$} & \close{$1.92240$} & $5.77869$\\
                        & NetGAN    &   $2485$ &  $10,138$ &  $932$ & \twoclose{$2,394$} & \twoclose{$128$} &  $0.15687$ & $-0.07384$ & $1.86148$ & \twoclose{$5.86039$}\\
                        & CELL &   $2708$ 
                         &  $10,556$ &  $521$ & $1,295$ & $97$ &  $0.07930$ & $-0.08226$ & \twoclose{$1.88642$} & \close{$6.03067$}\\
                         & DCSBM &   $2621$ 
                         &  $10,097$ &  $2,380$ & \close{$2,809$} & $237$ &  $0.06991$ & $-0.01709$ & $1.85581$ & $4.47947$\\
                         \midrule

\multirow{ 6}{*}{\rotatebox{90}{Wiki}}
& Real        &   $2405$ &  $12,761$ &  $23,817$ & $407,302$ & $263$ &  $0.37581$ &  $-0.07875$ & $1.54227$ & $ 3.65161$\\ 
                        & {\SaGess}-RW  &   $2348$ &  $12,763$ &  \close{$26,296$}  & \close{$466,120$} & $280$ & \twoclose{$0.38389$} &  \close{$-0.09282$} & \close{$1.54076$}& $3.53405$\\ 
                        & {\SaGess}-Ego &   $2275$ &  $12,763$ &  \twoclose{$26,897$}  & \twoclose{$506,620$} & $312$ & \close{$0.37530$} &  $-0.10494$ & $1.55261$ & $3.47834$\\ 
                         & NetGAN    &   $2405$ &  $11,596$ &  $10,701$ & $104,243$& $241$ &  $0.20179$ & \twoclose{$-0.10076$} & $1.52944$ & \twoclose{$3.60293$} \\
                         & CELL &   $2357$ 
                         &  $11,592$ &  $10,136$ & $ 121,014 $ & \twoclose{$258$} &  $0.22965$ & $-0.10769$ & \twoclose{$1.54027$} & \close{$3.62619$}\\ 
                         & DCSBM &   $2251$ 
                         &  $11,595$ &  $9,439$ & $ 167,655 $ & \close{$263$} &  $0.08492$ & $-0.01821$ & $1.54596$ & $3.45205$\\ \midrule

\multirow{ 7}{*}{\rotatebox{90}{Facebook}}
& Real        &   $1045$ &  $27,755$ &  $446,846$  & $34,098,662$& $1044$& $0.57579$ &  $-0.02543$ & $1.28698$ & $1.94911$\\ 
                        & {\SaGess}-Uni&   $1043$ &  $27,758$ &  \close{$429,428$} & \close{$35,261,545$} & \twoclose{$999$} &  \twoclose{$0.52098$} &  $-0.01607$ & \twoclose{$1.29003$} & \twoclose{$2.00800$} \\ 
                        & {\SaGess}-RW &   $1009$ &  $27,764$ &  \twoclose{$ 490,844$}  & $ 43,006,252$ & \close{$1001$} & \close{$0.56138$} &  \close{$-0.02266$} & $1.29398$ & \close{$1.96014$}\\
                        & {\SaGess}-Ego &   $1005$ &  $27,761$ &  $ 515,928$  & $45,421,130$ & $295$ & $0.43074$ &  $0.34074$ & $1.29381$ & $2.65926$\\
                         & NetGAN    &   $1045$ &  $27,755$ &  $262,574$  & $15,635,262$ & $849$ & $0.39773$ & \twoclose{$-0.01821$} & $1.27429$ & $2.13730$\\
                         & CELL &   $1045$ 
                         &  $27,755$ &  $250,968$ & $14,855,676$ & $474$ &  $0.30854$ & $0.12788$ & $1.27490$ & $ 2.38650$\\
                         & DCSBM &   $1041$ 
                         &  $27,092$ &  $339,448$ & \twoclose{$26,714,948$} & $733$ &  $0.37549$ & $0.07125$ & \close{$1.28845$} & $ 2.33021$\\\midrule

\multirow{ 7}{*}{\rotatebox{90}{SBM}}
& Real        &   $1600$ &  $73,312$ &  $344,574$  & $20,955,308$ & $155$& $0.15418$ &  $-0.00188$ & $3.58894$ & $2.08276$\\ 
                        & {\SaGess}-Uni&   $1600$ &  $73,313$ &  $\close{342,639}$ & $ \twoclose{21,054,221}$ & $164$ &  $0.14830$ &  $0.03572$ & $ 2.16732$ & $2.04884$ \\ 
                        & {\SaGess}-RW &   $1600$ &  $73,367$ &  $ 490,663$  & $34,384,218$ & $207$ & $0.20062$ &  $0.22337$ & $1.70495$ & $ 2.13963$\\
                        & {\SaGess}-Ego &   $1600$ &  $73,326$ &  $ 366,162$  & $22,979,792$ & $174$ & \close{$0.15622$} &  $0.07177$ & $1.89330$ & $2.06118$\\
                         & NetGAN    &   $1600$ &  $73,312$ &  $ 367,143$  & $ 22,775,320$ & $144$ & $0.16054$ & $0.19039$ & $2.39375$ & $2.12747$\\
                         & CELL &   $1600$ &  $73,312$ &  \twoclose{$341,351$} & \close{$20,783,575$} & \twoclose{$139$} &  $0.15186$ & \close{$-0.00156$} & \twoclose{$2.72805$} & \twoclose{$2.08060$}\\
                         & DCSBM &   $1600$ 
                         &  $73,357$ &  $353,934$ & $21,794,585$ & \close{$130$} &  \twoclose{$0.15671$} & \twoclose{$ -0.00343$} & \close{$2.79204$} & \close{$ 2.08274$}\\
                        \bottomrule
\end{tabular}
\end{adjustbox}

\end{table}

We evaluate our framework using three different experimental settings. In our first experiment, we evaluate how close the structural properties of the generated graphs from each of the methods are to the original graph. For evaluation we choose the set of standard metrics from Ref.~\cite{pmlr-v119-rendsburg20a} such as  
number of nodes  (ignoring isolated nodes), 
clustering coefficient, assortativity or triangle count.  
As some generation methods fix the number of nodes and others fix the number of edges, we report these numbers also but do not assess the methods on them.

To further explore our  approach, we additionally compare the effectiveness of each of our 
sampling schemes using the same approach. We produce and compare synthetic graphs for all data sets for the Random Walk (RW), and $2$-hop modified neighborhood (Ego) sampling schemes.
While for the Uniform (Unif) sampling scheme, we compare on EuCore, Facebook and SBM, noting that this sampling scheme struggles on the remaining data sets, due to their sparsity.  

In essence, as the induced subgraphs in the training set from the uniform sampling tend to have very few edges (less than the nodes of the subgraph), our model will struggle to find the right amount of edges to match the edge count of the initial graph. 

The second experiment consists of a utility test. Indeed, we want to evaluate the usefulness
of the synthetic data set produced. Hence, we propose to train a Variational Graph Auto-Encoder for link prediction, the GVAE is trained on the generated synthetic data. 
Once trained, we obtain a latent space from the synthetic data set on the set of nodes belonging to the real data set. The  evaluation is the following: we compute the probability of each possible edge using the GVAE encodings and evaluate them on the $90$\% of the edges of the real graph. This can also be interpreted as a test on how efficient the synthetic data set can be, when employed for a data augmentation task.

Finally, we also explore the ability of our method to produce meaningful 
smaller graph samples.
Indeed, it might not be needed to generate a graph of the same size as the initial graph $G$. Our goal is to demonstrate that our model can also generate smaller scale synthetic graphs in terms of edges that are structurally similar to the initial graph. We generate graphs of sizes ranging from $10\%$ to $100\%$ of the number of edges in the initial graph, and observe the computed metrics convergences to the right values in \Cref{fig:progressive}.
All experimental hyperparameters, as well as extended experiments can be found in the supplementary. 

\section{Results}
\label{sec:results}

Throughout our experiments, we have shown that {\SaGess} is a very powerful tool which can create high quality synthetic graphs. 
This is demonstrated in our first experiment, where the statistics on the synthetic graphs (\Cref{tab:table1}) we produce, perform well in comparison to the benchmark models, across all the benchmark data sets. In particular, the {\SaGess}-RW method 
performs well  in the first four data sets, although struggles to some extent in the synthetic SBM data set. 
It is worth noticing as well that we tend to generate local structures, such as triangles, easier than the other baselines. This is due to the fact that {\DiGress} is very efficient at sampling motifs, and the subgraphs used for training contain many of them due to the nature of the sampling, which samples subgraphs with interdependencies between the edges. It is important to point out that even if we train on local substructures via the subgraphs, we manage to accurately obtain global metrics like characteristic path length or clustering coefficient.

In our second experiment, we present the results in 
 \Cref{tab:table2} where we exhibit the utility of our method that overall performs very consistently compared to the other methods. We also notice that while we are close to other methods in terms of statistics on some data sets, the utility performance on the link prediction is somewhat more diverse. For example, while the DCSBM method performs well in terms of graph statistics on the synthetic SBM data set (see \Cref{tab:table1}), we still outperform it by a significant factor in the link prediction task. 

\begin{figure}[t]
    \centering
    \includegraphics[width = \textwidth]{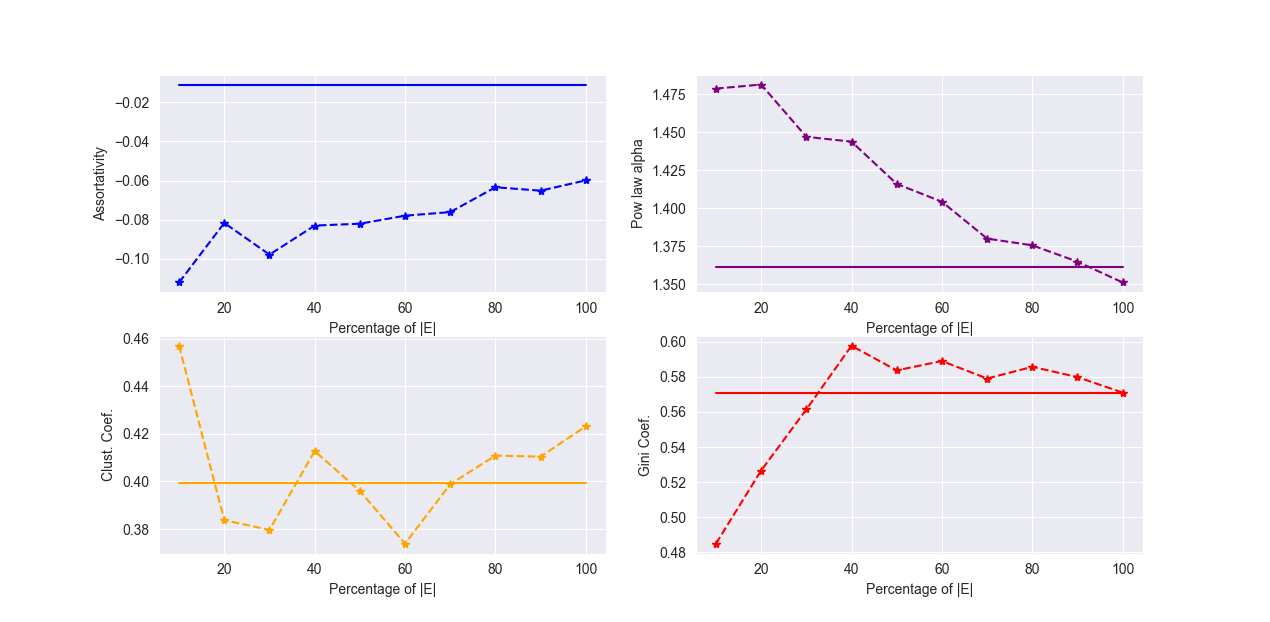}
    \caption{Graphs statistics as a function of percentage of sampled edges 
    from the EuCore data set using the {\SaGess}-RW model.
    Note, $y$-axis is scaled to data, however, the amount of variation of the statistics is relatively small. }    
    \label{fig:progressive}
\end{figure}

Finally, we can also conclude from \Cref{fig:progressive} that our model can generate graphs 
with a substantially smaller number of edges, while maintaining reasonable approximations of the graph statistics.  
For example, smaller samples have connectivity and structural properties that are similar to the initial graph even with $40\%$ of its edges. 
This is not only an interesting feature of our sampling scheme/generator, it is also potentially useful from a computational perspective, when one needs to deploy this methodology on machines with limited resources or shorter time frames, 
especially as the complexity of graph operations often depends on the number of edges as well as nodes.

\begin{table} 

\caption{\footnotesize Link prediction using
GVAE on the EuCore dataset, trained on the synthetic, tested on real.}
\label{tab:table2}
    \centering
    \footnotesize
    \begin{tabular}{ccccccc}
    \toprule
Method  & Sampling & EuCore & Cora & Wiki & Facebook & SBM\\ 
\midrule
Unif-{\SaGess} & AUC & $0.8463$ & - & - & $0.8601$ & $0.6801$\\
               & AP & $0.8330$ & - & - & $0.8415$ & $0.6439$ \\ \midrule
RW-{\SaGess} & AUC & $0.8455$ & $\mathbf{0.8952}$ & $\mathbf{0.8844}$ & $\mathbf{0.8993}$ & $0.6921$ \\
               & AP & $0.8340$ & $\mathbf{0.8948}$ & $\mathbf{0.8940}$ & $\mathbf{0.8862}$ & $0.6314$ \\ \midrule
Ego-{\SaGess} & AUC & $0.8463$ & $0.8875$ & $0.8707$ & $0.8889$ & $0.6991$ \\
               & AP & $0.8330$ & $0.8899$ & $0.8856$ & $0.8773$ & $0.6445$ \\ \midrule
NetGAN & AUC & $0.8123$ & $0.5099$ & $0.5613$ & $0.8703$ & $0.6978$ \\
               & AP & $0.8156$ & $0.5101$ & $0.5830$ & $0.8499$ & $0.6407$ \\ \midrule

CELL & AUC & $\mathbf{0.8566}$ & $0.7903$ & $0.8408$ & $0.8714$ & $\mathbf{0.7001}$ \\
               & AP & $\mathbf{0.8456}$ & $0.8056$ & $0.8637$ & $0.8467$ & $\mathbf{0.6479}$ \\ \midrule
DCSBM & AUC & $0.6555$ & $0.5096$ & $0.4965$ & $0.4978$ & $0.5378$ \\
               & AP & $0.6383$ & $0.5036$ & $0.5113$ & $0.4976$ & $0.5261$ \\   

    \bottomrule
\end{tabular}
\end{table}

\section{Discussion and Conclusion}
\label{sec:concl}

In this paper, we proposed {\SaGess}, a sampling-based denoising diffusion probabilistic model, based on {\DiGress}. It is presented as a framework that is able to build a training data set from a single graph observation, and use it to generate a synthetic graph from a collection of generated synthetic subgraphs. We have shown throughout a variety of experiments that our method outperforms state-of-the-art methods on several real world data sets, not only by considering standard graph statistics, but also by showcasing the adaptability and utility of our framework. 

However, this does not come without cost. One limitation of our method is that it still scales poorly in terms of memory with the number of nodes in the initial graph. This is due to the one-hot-encoding of the graph node ids passed in {\DiGress}. On the same note, as the number of nodes increases, the number of subgraph samples needed to train our model also increases, adding to the time complexity of {\DiGress}. In essence the memory complexity depends on the size of a subsample, resulting in $O(k^2+kn)$ where $k$ is the size of the subgraph. Additionally, while this method could trivially deal with edge attributes as {\DiGress} does, it cannot handle node attributes since they are already taken by the node identification step.

An advantage of our approach is that it can be extended to address other related problems. 
While this method has been designed to generate graphs from a single sample, it can be extended to data sets with multiple graphs in the case where the nodes are identifiable, by simply designing a sampling scheme which first samples the graph and then samples the subgraph using an appropriate scheme, although we leave this avenue for future work. It will also be left as future work to investigate possible applications on signed networks or even time dependent edges, as one can add time as an edge attribute. This could lead to a more complex generation scheme that could, for instance, bear an auto-regressive module to evaluate time dependence on the newly encountered edges.  
We would also be eager to extend our framework to other graph generation methods that handle node attributes to generalize the ability of models to train on a single observation.

Finally, some caution is advised. Inference based on synthetic data particularly in sensitive applications such as health or risk analysis should use a number of different synthetic data generators before reaching a conclusion. Moreover, vigilance is advised to detect malicious applications of synthetic data generation, such as passing off fake data as real.

\bibliographystyle{plain}
\bibliography{biblio.bib}

\begin{thebibliography}{10}

\bibitem{RevModPhys.74.47}
R\'eka Albert and Albert-L\'aszl\'o Barab\'asi.
\newblock Statistical mechanics of complex networks.
\newblock {\em Rev. Mod. Phys.}, 74:47--97, Jan 2002.

\bibitem{ali2016comparison}
Waqar Ali, Anatol~E Wegner, Robert~E Gaunt, Charlotte~M Deane, and Gesine
  Reinert.
\newblock Comparison of large networks with sub-sampling strategies.
\newblock {\em Scientific reports}, 6(1):28955, 2016.

\bibitem{beaulieu2019privacy}
Brett~K Beaulieu-Jones, Zhiwei~Steven Wu, Chris Williams, Ran Lee, Sanjeev~P
  Bhavnani, James~Brian Byrd, and Casey~S Greene.
\newblock Privacy-preserving generative deep neural networks support clinical
  data sharing.
\newblock {\em Circulation: Cardiovascular Quality and Outcomes},
  12(7):e005122, 2019.

\bibitem{bojchevski2018netgan}
Aleksandar Bojchevski, Oleksandr Shchur, Daniel Z{\"u}gner, and Stephan
  G{\"u}nnemann.
\newblock Netgan: Generating graphs via random walks.
\newblock In {\em International conference on machine learning}, pages
  610--619. PMLR, 2018.

\bibitem{erdHos1960evolution}
Paul Erd{\H{o}}s and Alfr{\'e}d R{\'e}nyi.
\newblock On the evolution of random graphs.
\newblock {\em Publ. Math. Inst. Hung. Acad. Sci}, 5(1):17--60, 1960.

\bibitem{faez2021deep}
Faezeh Faez, Yassaman Ommi, Mahdieh~Soleymani Baghshah, and Hamid~R Rabiee.
\newblock Deep graph generators: A survey.
\newblock {\em IEEE Access}, 9:106675--106702, 2021.

\bibitem{guo2022systematic}
Xiaojie Guo and Liang Zhao.
\newblock A systematic survey on deep generative models for graph generation.
\newblock {\em IEEE Transactions on Pattern Analysis and Machine Intelligence},
  2022.

\bibitem{haefeli2022diffusion}
Kilian~Konstantin Haefeli, Karolis Martinkus, Nathana{\"e}l Perraudin, and
  Roger Wattenhofer.
\newblock Diffusion models for graphs benefit from discrete state spaces.
\newblock {\em arXiv preprint arXiv:2210.01549}, 2022.

\bibitem{ho2020denoising}
Jonathan Ho, Ajay Jain, and Pieter Abbeel.
\newblock Denoising diffusion probabilistic models.
\newblock {\em Advances in Neural Information Processing Systems},
  33:6840--6851, 2020.

\bibitem{HOLLAND1983109}
Paul~W. Holland, Kathryn~Blackmond Laskey, and Samuel Leinhardt.
\newblock Stochastic blockmodels: First steps.
\newblock {\em Social Networks}, 5(2):109--137, 1983.

\bibitem{karrer2011stochastic}
Brian Karrer and Mark~EJ Newman.
\newblock Stochastic blockmodels and community structure in networks.
\newblock {\em Physical review E}, 83(1):016107, 2011.

\bibitem{liao2019gran}
Renjie Liao, Yujia Li, Yang Song, Shenlong Wang, Charlie Nash, William~L.
  Hamilton, David Duvenaud, Raquel Urtasun, and Richard Zemel.
\newblock Efficient graph generation with graph recurrent attention networks.
\newblock In {\em NeurIPS}, 2019.

\bibitem{facebook}
Julian McAuley and Jure Leskovec.
\newblock Learning to discover social circles in ego networks.
\newblock In {\em Proceedings of the 25th International Conference on Neural
  Information Processing Systems - Volume 1}, NIPS'12, page 539–547, Red
  Hook, NY, USA, 2012. Curran Associates Inc.

\bibitem{mcgregor_et_al:LIPIcs.ICALP.2022.96}
Andrew McGregor and Rik Sengupta.
\newblock {Graph Reconstruction from Random Subgraphs}.
\newblock In Miko{\l}aj Boja\'{n}czyk, Emanuela Merelli, and David~P. Woodruff,
  editors, {\em 49th International Colloquium on Automata, Languages, and
  Programming (ICALP 2022)}, volume 229 of {\em Leibniz International
  Proceedings in Informatics (LIPIcs)}, pages 96:1--96:18, Dagstuhl, Germany,
  2022. Schloss Dagstuhl -- Leibniz-Zentrum f{\"u}r Informatik.

\bibitem{pmlr-v119-rendsburg20a}
Luca Rendsburg, Holger Heidrich, and Ulrike~Von Luxburg.
\newblock {N}et{GAN} without {GAN}: From random walks to low-rank
  approximations.
\newblock In Hal~Daumé III and Aarti Singh, editors, {\em Proceedings of the
  37th International Conference on Machine Learning}, volume 119 of {\em
  Proceedings of Machine Learning Research}, pages 8073--8082. PMLR, 13--18 Jul
  2020.

\bibitem{ROBINS2007173}
Garry Robins, Pip Pattison, Yuval Kalish, and Dean Lusher.
\newblock An introduction to exponential random graph (p*) models for social
  networks.
\newblock {\em Social Networks}, 29(2):173--191, 2007.
\newblock Special Section: Advances in Exponential Random Graph (p*) Models.

\bibitem{schneuing2022structure}
Arne Schneuing, Yuanqi Du, Charles Harris, Arian Jamasb, Ilia Igashov, Weitao
  Du, Tom Blundell, Pietro Li{\'o}, Carla Gomes, Max Welling, et~al.
\newblock Structure-based drug design with equivariant diffusion models.
\newblock {\em arXiv preprint arXiv:2210.13695}, 2022.

\bibitem{simonovsky2018graphvae}
Martin Simonovsky and Nikos Komodakis.
\newblock Graphvae: Towards generation of small graphs using variational
  autoencoders.
\newblock In {\em Artificial Neural Networks and Machine Learning--ICANN 2018:
  27th International Conference on Artificial Neural Networks, Rhodes, Greece,
  October 4-7, 2018, Proceedings, Part I 27}, pages 412--422. Springer, 2018.

\bibitem{sohl2015deep}
Jascha Sohl-Dickstein, Eric Weiss, Niru Maheswaranathan, and Surya Ganguli.
\newblock Deep unsupervised learning using nonequilibrium thermodynamics.
\newblock In {\em International Conference on Machine Learning}, pages
  2256--2265. PMLR, 2015.

\bibitem{van2021decaf}
Boris van Breugel, Trent Kyono, Jeroen Berrevoets, and Mihaela van~der Schaar.
\newblock Decaf: Generating fair synthetic data using causally-aware generative
  networks.
\newblock {\em Advances in Neural Information Processing Systems},
  34:22221--22233, 2021.

\bibitem{vignac2022digress}
Clement Vignac, Igor Krawczuk, Antoine Siraudin, Bohan Wang, Volkan Cevher, and
  Pascal Frossard.
\newblock Digress: Discrete denoising diffusion for graph generation.
\newblock {\em arXiv preprint arXiv:2209.14734}, 2022.

\bibitem{wong2016understanding}
Sebastien~C Wong, Adam Gatt, Victor Stamatescu, and Mark~D McDonnell.
\newblock Understanding data augmentation for classification: when to warp?
\newblock In {\em 2016 international conference on digital image computing:
  techniques and applications (DICTA)}, pages 1--6. IEEE, 2016.

\bibitem{xu2022geodiff}
Minkai Xu, Lantao Yu, Yang Song, Chence Shi, Stefano Ermon, and Jian Tang.
\newblock Geodiff: A geometric diffusion model for molecular conformation
  generation.
\newblock {\em arXiv preprint arXiv:2203.02923}, 2022.

\bibitem{yang2020scaling}
Renchi Yang, Jieming Shi, Xiaokui Xiao, Yin Yang, Juncheng Liu, and Sourav~S
  Bhowmick.
\newblock Scaling attributed network embedding to massive graphs.
\newblock {\em arXiv preprint arXiv:2009.00826}, 2020.

\bibitem{yang2016revisiting}
Zhilin Yang, William~W. Cohen, and Ruslan Salakhutdinov.
\newblock Revisiting semi-supervised learning with graph embeddings, 2016.

\bibitem{eucore}
Hao Yin, Austin~R. Benson, Jure Leskovec, and David~F. Gleich.
\newblock Local higher-order graph clustering.
\newblock In {\em Proceedings of the 23rd ACM SIGKDD International Conference
  on Knowledge Discovery and Data Mining}, KDD '17, page 555–564, New York,
  NY, USA, 2017. Association for Computing Machinery.

\bibitem{you2018graphrnn}
Jiaxuan You, Rex Ying, Xiang Ren, William Hamilton, and Jure Leskovec.
\newblock Graph{RNN}: Generating realistic graphs with deep auto-regressive
  models.
\newblock In {\em International conference on machine learning}, pages
  5708--5717. PMLR, 2018.

\end{thebibliography}

\end{document}